\documentclass[10pt,twocolumn,letterpaper]{article}

\usepackage{wacv}
\usepackage{times}
\usepackage{epsfig}
\usepackage{graphicx}
\usepackage{amsmath}
\usepackage{amssymb}
\usepackage{multirow}
\usepackage{float}
\usepackage{url}

\wacvfinalcopy 

\def\wacvPaperID{133} 

\ifwacvfinal\fi
\title{Robust Point Light Source Estimation Using Differentiable Rendering}

\author{Gr\'egoire Nieto \\
Technicolor\\
{\tt\small gregoire.nieto@gmail.com}
\and
Salma Jiddi \\
Technicolor\\
{\tt\small salma.jiddi@technicolor.com}
\and
Philippe Robert \\
Technicolor\\
{\tt\small philippe.robert@technicolor.com}
}

\DeclareMathOperator*{\argmin}{arg\,min}

\begin{document}

\maketitle
\ifwacvfinal\thispagestyle{empty}\fi

\begin{figure*}[t]
\begin{center}
\begin{tabular}{cc}
\includegraphics[width=0.45\textwidth]{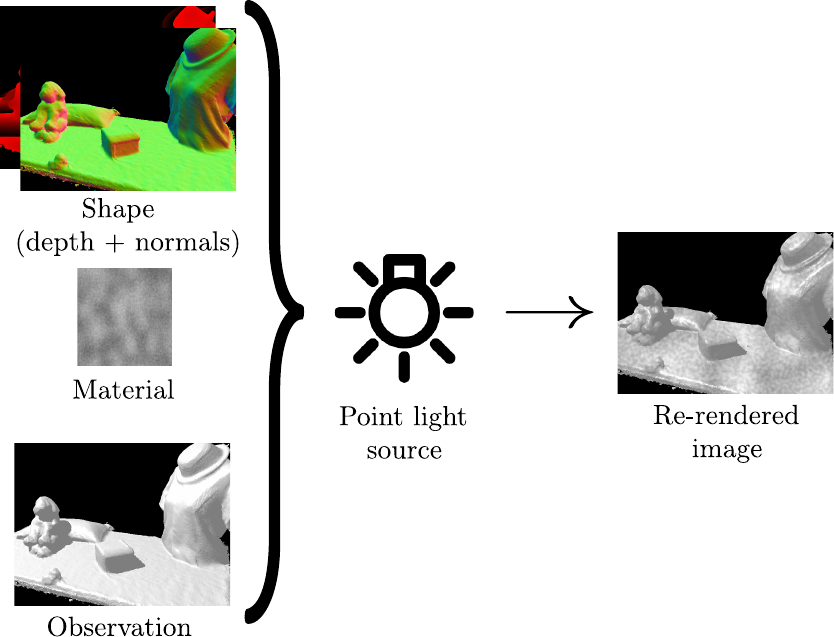} &
\includegraphics[width=0.45\textwidth]{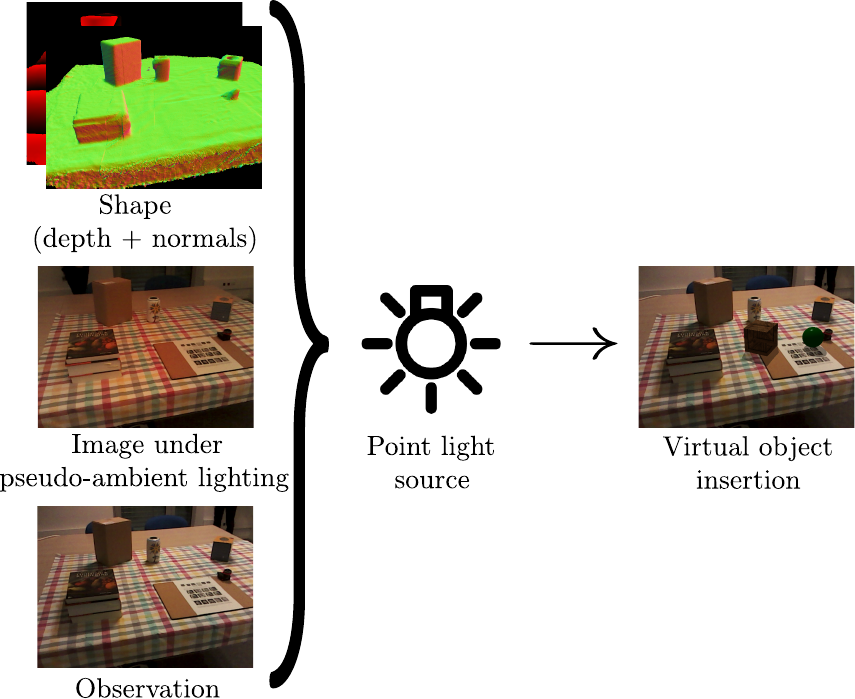} \\
(a) & (b)
\end{tabular}
\end{center}
\caption{Our method takes as input the shape (depth and normals), the material properties (reflectance), and given an observed image it estimates the illumination modeled as a point light source. (a) To test its robustness to incorrect material estimation, we intentionally provide erroneous noisy material and re-render a synthetic scene with the estimated illumination. (b) An image captured under pseudo-ambient lighting is given as input to recover illumination of real scenes. The estimated light source is used to insert virtual objects with consistent lighting for mixed reality applications (cube and sphere).}
\label{fig:teaser}
\end{figure*}

\begin{abstract}
Illumination estimation is often used in mixed reality to re-render a scene from another point of view, to change the color/texture of an object, or to insert a virtual object consistently lit into a real video or photograph. Specifically, the estimation of a point light source is required for the shadows cast by the inserted object to be consistent with the real scene. We tackle the problem of illumination retrieval given an RGBD image of the scene as an inverse problem: we aim to find the illumination that minimizes the photometric error between the rendered image and the observation. In particular we propose a novel differentiable renderer based on the Blinn-Phong model with cast shadows. We compare our differentiable renderer to state-of-the-art methods and demonstrate its robustness to an incorrect reflectance estimation.
\end{abstract}

\section{Introduction}
\label{sec:intro}

{\bf Inverse rendering} often refers to estimating missing parameters of a scene, given a rendered image. Among geometry and material, illumination is a crucial aspect in the production of an image. Knowing such element could be used to re-render the scene from another point of view, to change the color/texture of an object, or to insert a virtual object consistently lit into a real video or photograph (figure \ref{fig:teaser} (b)). From a mixed reality perspective, a use case would be to capture the scene from a fixed viewpoint with an RGBD sensor and forward the data to a server that would estimate the light source at interactive frame rate. Then the virtual object would be rendered with consistent shadows and shading on an Augmented Reality (AR) device such as a tablet or a glass-type Head-Mounted Display (HMD). For the shadows cast by the virtual object to be consistent with the real lighting, the estimation of non-distant punctual light sources is needed, as opposed to the estimation of environment maps or directional (infinitely distant) light sources. 

We tackle the problem of illumination retrieval given an RGBD image of the scene as an {\bf inverse problem}: "the approach tries to reverse-engineer the physical process that produced an image of the world"~\cite{loper_opendr:_2014}. Instead of directly finding a candidate for light sources, we assume that the illumination is known and we maximize the likelihood to get the observed 2D image. It requires rendering --- how illumination interacts with the material and the geometry to produce an image --- to be completely differentiable. Then we are able to propagate the photometric error between the reconstructed image and the observation back to the lighting parameters. That is why we say  our method uses {\bf differentiable rendering}. Given the Jacobian matrix of the rendered image with respect to the lighting parameters, the chain rule enables us to iteratively adjust the lighting parameters that eventually converge to an estimate. In the machine learning literature, such algorithm is called {\bf gradient back-propagation} (here the term "gradient" defines the derivative of the loss with respect to the lighting parameters).

In the domain of illumination retrieval, we can address some criticisms to state-of-the-art methods. Indeed, most of them do not make use of cast shadows when estimating the illumination \cite{jacobs2006classification, nishino_determining_2001, hara_light_2005, jiddi_reflectance_2016, Jiddi18a, lopez-moreno_multiple_2013, karsch_rendering_2011, boom2013point, neverova_lighting_2012}, and when they do \cite{jiddi_illumination_2017} they rarely take diffuse and specular reflections into account. Moreover these algorithms often rely on explicit shadow detection. Methods that use a complex reflection model that handles specularities often compute shadows with non-differentiable shadow mapping, to perform light source matching or a discrete form of optimization. On the other hand we propose to apply our differentiable renderer to {\bf continuous} optimization, which means the search space of light source positions is not constrained to discrete values. We solve the optimization problem by inverse rendering with the {\bf Blinn-Phong model}, {\bf including shadow casting}, so that both specularities and cast shadows are implicitly used without any explicit diffuse/specular separation nor shadow detection. We show that appending a shadow term increases the robustness of the estimation in the case of an imperfect reflectance estimation (figure \ref{fig:teaser} (a)), which is the common practical case.

Our differentiable renderer can also be applied to improve the training of neural networks. Recent progress in machine learning enables the estimation of illumination from a single image via a simple neural network. Such network performs better when it is trained in a self-supervised manner. It means the same training example --- the observed image --- is used as input and ground truth at the same time. It is passed to the network to train that turns it into an estimate of the illumination, then an image is reconstructed knowing the geometry and the materials. The photometric error between the input image and the reconstructed image is back-propagated to adjust the weights of the neural network. To the best of our knowledge we are the first to propose an implementation of a {\bf renderer with differentiable cast shadows} that could process simultaneously batches of images and be integrated into an {\bf unsupervised deep learning architecture}.

To sum up, our contributions are listed as follow: 
\begin{itemize}
    \item a completely differentiable rendering pipeline that produces Blinn-Phong~\cite{phong_illumination_1975-1, blinn_models_1977} shading with cast shadows;
    \item a novel point light source estimation based on optimization that cleverly makes use of the presence of shadows to compensate for an imprecise reflectance estimation;
    \item an efficient batch implementation designed to train unsupervised deep learning architectures~\cite{georgoulis_reflectance_2018, wang_joint_2017, janner_self-supervised_2017, rematas_deep_2016, georgoulis_delight-net:_2016}.
\end{itemize}

\section{Related Work}
\label{sec:related}

A classification of illumination estimation methods is proposed by Jacobs and Loscos \cite{jacobs2006classification}. Existing approaches are numerous, thus in this section we especially focus on methods that cope with cast shadows or specularities. Nishino \etal \cite{nishino_determining_2001}, Hara \etal \cite{hara_light_2005} and Jiddi \etal \cite{jiddi_reflectance_2016, Jiddi18a} aim to recover the illumination from specular profiles, but assume that at least one specular peak is visible. Lopez-Moreno \etal \cite{lopez-moreno_multiple_2013} estimate multiple point light sources: albedo and highlights are first removed from the object; then its silhouette is used to infer the light source position in screen-space; finally its interior is used to estimate the position in world space. However convexity near the silhouette is assumed for normal computation.

\paragraph{Light source candidates} A way to cope with cast shadows is to model illumination as a set of candidates, either point sources \cite{jiddi_illumination_2017} or directional sources \cite{sato_illumination_1999}. Sato \etal \cite{sato_illumination_1999} propose to estimate the illumination distribution by an adaptive sampling of the directional sources. They implicitly make use of shadows by incorporating a shadow term in their Bidirectional Reflectance Distribution Function (BRDF), computed from the geometry and the potential directional light candidates. In \cite{jiddi_illumination_2017}, Jiddi \etal approximate a set of point light sources equally distributed around the scene, and select the candidates whose cast shadows correlate with a binary mask of shadows preliminary detected in the image. Our method differs from \cite{jiddi_illumination_2017} and \cite{sato_illumination_1999} in that we perform a continuous optimization, and we do not approximate a set of candidates. Hence there is no need to compute a set of shadow maps prior to the optimization: a shadow map is rendered at each iteration given the current illumination estimate.

\paragraph{Optimization-based methods} The closest work to ours is Neverova \etal's \cite{neverova_lighting_2012}: it is an optimization-based method with Phong rendering to estimate a point light source from RGBD images. It is inspired by the work of Karsch \etal \cite{karsch_rendering_2011} that only deals with diffuse objects and a very coarse geometry. Boom \etal \cite{boom2013point} also deal with Lambertian surfaces, although their optimization scheme is close to ours. Alike \cite{neverova_lighting_2012}, Mashita \etal \cite{mashita_parallel_2013-1} use a Phong model but focus on the effect of using multiple views. All these methods \cite{neverova_lighting_2012, mashita_parallel_2013-1, boom2013point, karsch_rendering_2011} do not take cast shadows into account. Neverova \etal's work \cite{neverova_lighting_2012} also relies on the separation between specular and diffuse components whereas our method implicitly includes both components in the optimization via the rendering equation, without explicit separation. Moreover we do not explicitly segment images or extract any surface properties like in~\cite{karaoglu_point_2017}. Our only model is the rendering model and we do not treat differently some parts of the image whether they are glossy, matte, highlighted, curved, etc.

\paragraph{Supervised learning} From a machine learning perspective, state-of-the-art techniques aim to estimate illumination (either indoor \cite{gardner_learning_2017-1} or outdoor \cite{hold-geoffroy_deep_2017}) by learning the weights of a neural network. To train such network, a supervised strategy is often adopted: the loss function is the error between some ground truth illumination and the illumination computed by the network fed with an example image. Rematas \etal \cite{rematas_deep_2016} learn to estimate an intermediate representation that mixes illumination and a single material, called a "reflectance map". In \cite{georgoulis_reflectance_2018} this architecture is combined with a CNN that decomposes the reflectance map into an environment map and a single material. In \cite{wang_joint_2017} a parametric model of illumination and material is fitted to a reflectance map, via a light transport operator that is approximated by two neural networks preliminary trained on synthetic data. Mandl \etal \cite{mandl_learning_2017-1} try to estimate the illumination only, by training a CNN for every camera pose sampled around an object used as a light probe.

\paragraph{Unsupervised learning} Recent work in neural networks has demonstrated the superiority of unsupervised strategies, for face reconstruction for example \cite{tewari_mofa:_2017, tewari_self-supervised_2017, sengupta_sfsnet:_2017}. At each iteration a differentiable renderer reconstructs an image from the current illumination estimation and compares it to the input example image. Janner \etal \cite{janner_self-supervised_2017} propose to train a renderer that consists of an encoder/decoder to produce a diffuse shading given an illumination estimation and a normal map. However, even with relevant training data, such architecture is unable to correctly mimic the shadow formation process. A differentiable renderer that performs shadow casting is yet required to produce a prediction that is as close as possible to what we can expect from indoor scenes. We believe our differentiable rendering module could highly benefit the deep learning community for all kind of inverse rendering applications.

\section{Our Approach}
\label{sec:approach}

\subsection{Background}

We model illumination as a punctual light source, parametrized by its 3D location ${\bf L}_i $ and a scalar intensity ${\bf I}_{L,i}$. Note that the method is not constrained by the use of a single light source, as long as all lights $i$ are punctual, in order to cope with realistic shadow casting. ${\bf I({\bf x})}$ is the intensity of the rendered image at point ${\bf x}$; to simplify we note this intensity ${\bf I}$. We choose to represent the geometry of the scene by ${\bf G} = ({\bf X}, {\bf N}, {\bf C})$, with ${\bf X}$ being a 3D point cloud, ${\bf N}$ the normals, and ${\bf C}$ the camera pose. Representing the geometry by an oriented point cloud is purely arbitrary. In practice, other data representations can be adopted: depth maps can easily be converted into a point cloud and if our implementation does not handle rasterization yet, it is only a matter of technical convenience. ${\bf G}$ also contains the camera pose, whose parameters can be estimated by any means, but that is not in the scope of the present work. Images are represented by vectors of size the number of pixels; all channels, if plural, are treated independently. The multiplication marked as a dot $.$ denotes the element-wise multiplication. The dot product or scalar product is denoted by the multiplication with the transposed vector. 

\begin{figure}
\begin{center}
\includegraphics[width=0.90\columnwidth]{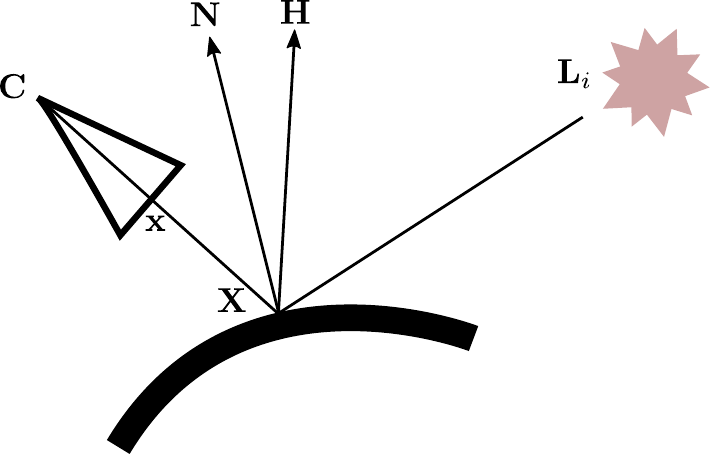}
\end{center}
\caption{Main vectors used in the Blinn-Phong model}
\label{fig:phong}
\end{figure}

\begin{table}
\begin{center}
\begin{tabular}{|l|c|}
\hline
${\bf I}_a$ & ambient term \\
\hline
${\bf I}_d$ & diffuse term \\
\hline
${\bf I}_s$ & specular term \\
\hline
${\bf L}_a$ & ambient illumination \\
\hline
${\bf L}_i$ & position of the light $i$ \\
\hline
${\bf I}_{L,i}$ & intensity of the light $i$ \\
\hline
${\bf S}_i$ & shadow term of the light $i$ \\
\hline
${\bf X}$ & 3D point on the surface \\
\hline
${\bf x}$ & 2D point on the screen \\
\hline
${\bf C}$ & camera center \\
\hline
${\bf N}$ & normal to the surface \\
\hline
${\bf H}$ & halfway vector \\
\hline
${\bf k}_d$ & diffuse reflectance \\
\hline
${\bf k}_s$ & specular reflectance \\
\hline
$\alpha$ & shininess \\
\hline
\end{tabular}
\end{center}
\caption{List of terms used by the Blinn-Phong reflection model.}%
\label{table:phong}
\end{table}

The Blinn-Phong model claims that the image intensity at point ${\bf x}$ (figure \ref{fig:phong}) is the sum of three terms ${\bf I} = {\bf I}_a + {\bf I}_d + {\bf I}_s$. All terms are listed in table~\ref{table:phong}. The ambient term is the multiplication of the ambient illumination --- due to an infinite number of inter-reflections --- with the diffuse reflectance: ${\bf I}_a = {\bf k}_d.{\bf L}_a$. Both specular and diffuse terms depend on the shadow term ${\bf S}_i$, that equals 1 when the light $i$ illuminates the point ${\bf X}$, and 0 otherwise. The diffuse term ${\bf I}_d$ describes the illumination of a Lambertian surface, \ie a surface whose BRDF is isotropic. For a 3D point ${\bf X}$ on the surface to render that is directly lit by the light source, --- for which the shadow term ${\bf S}_i$ equals one --- it is proportional to the scalar product of the normal ${\bf N}$ with the light source vector ${\bf L}_i - {\bf X}$: ${\bf I}_d = {\bf k}_d.\sum_i {\bf S}_i.({\bf L}_i - {\bf X})^t{\bf N}.{\bf I}_{L,i}$. The specular term ${\bf I}_s$ is a factor of the scalar product between the normal ${\bf N}$ and the halfway vector ${\bf H}$ between the viewer $({\bf C} - {\bf X})$ and the light source vector $({\bf L}_i - {\bf X})$ (see figure~\ref{fig:phong}): ${\bf I}_s = {\bf k}_s.\sum_i {\bf S}_i.({\bf H}^t{\bf N})^\alpha.{\bf I}_{L,i}$. Finally the intensity of the image can be written as follow:
\begin{equation}
\label{eq:phong}
    {\bf I} = {\bf k}_d.{\bf L}_a + \sum_i {\bf S}_i.{\bf I}_{L,i}.\left({\bf k}_d.({\bf L}_i - {\bf X})^t{\bf N} + {\bf k}_s.({\bf H}^t{\bf N})^\alpha\right).
\end{equation}

\subsection{Light Source Estimation}

\begin{figure}
\begin{center}
\includegraphics[width=0.90\columnwidth]{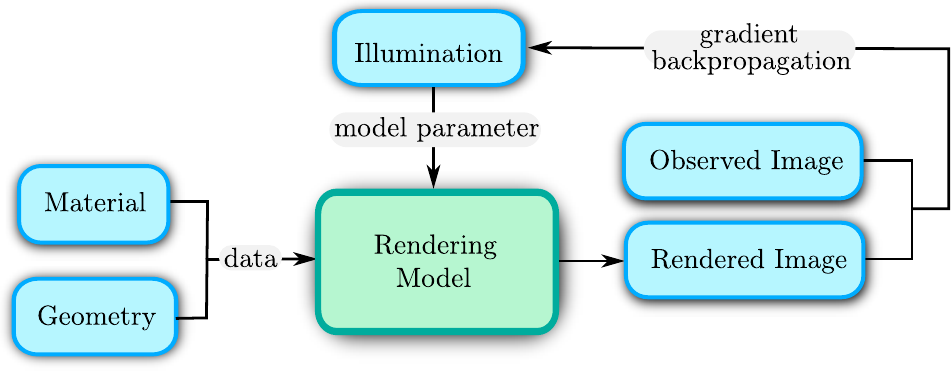}
\end{center}
\caption{Overview of our light source estimation procedure. }
\label{fig:model}
\end{figure}

Our approach to estimate a light source via optimization (see figure \ref{fig:model} for an overview) takes an observed image ${\bf I^*}$ as input. We assume the geometry of the scene ${\bf G} = ({\bf X}, {\bf N}, {\bf C})$ and the materials ${\bf M} = ({\bf k}_d, {\bf k}_s, \alpha)$ to be known, so that our Blinn-Phong model (\ref{eq:phong}) is only parametrized by the illumination ${\bf L}$. As mentioned before, we opt for an inverse approach, that takes advantage of our differentiable rendering pipeline. Rendering is combining ${\bf G}$, ${\bf M}$, and ${\bf L}$ to produce a 2D image ${\bf I}$. Given the imperfection of the chosen rendering model and the geometry and illumination estimates, the produced image ${\bf I}({\bf G}, {\bf M}, {\bf L})$ is different from the observed data ${\bf I^*}$. A good estimation of ${\bf L}$ must minimize the photometric error ${\bf E}$, \ie the $L^2$-norm of the difference between ${\bf I}$ and ${\bf I^*}$:
\begin{equation}
\label{eq:energy1}
    {\bf \hat{L}} = \argmin_{\bf L} {\bf E} = \argmin_{\bf L} \| {\bf I}({\bf G}, {\bf M}, {\bf L}) - {\bf I^*} \|^2.
\end{equation}
To minimize such energy (\ref{eq:energy1}), an iterative scheme is adopted, for example with a gradient descent algorithm. The light source is first initialized somewhere. Then at each iteration, an image is rendered given the current light source estimation, with shadows and specularities potentially. The rendered image is compared to the observed image and the error is back-propagated to adjust the illumination parameters. The derivative of the energy ${\bf E}$ with respect to the illumination parameters is $\partial {\bf E}/ \partial {\bf L} = ({\bf I} - {\bf I^*})^t \partial {\bf I}/\partial {\bf L}$.
It requires the computation of the Jacobian matrix $\partial {\bf I}/\partial {\bf L}$, which is detailed in the next section.

\subsection{Differentiable Rendering}

In this section we detail the derivation of the rendered image~(\ref{eq:phong}) with respect to the illumination parameters. These parameters are gathered all together in a single parameter vector ${\bf L} = ({\bf L}_i, {\bf I}_{L,i})_i$. To find the Jacobian matrix $\partial {\bf I}/\partial {\bf L}$, we choose to represent the image formation model as an acyclic graph, as it is commonly done in the neural network literature. An acyclic graph is a tree-like graph with no loop. Nodes are intermediate variables, linked by edges that represent transformations. The leaves of the graph are the input variables, that can be the data (geometry) or the model parameters (illumination); all paths converge to the root, which is the output (rendered image). A path from a leaf to the root is a composition of transformations, whose derivative is given by the chain rule. If we guarantee that each node produces a differentiable transformation, the chain rule assures that the derivative of the root with respect to the leaves exists. In other words we have to make sure there is in our graph at least one way from ${\bf L}$ to ${\bf E}$ that propagates the gradients.

\begin{figure}
\begin{center}
\includegraphics[width=0.90\columnwidth]{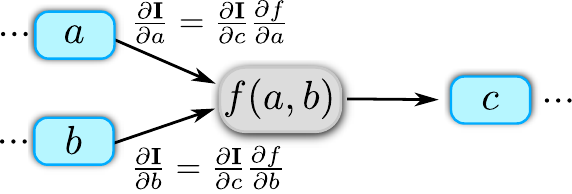}
\end{center}
\caption{Graph node and differentiability.}
\label{fig:nodeGraph}
\end{figure}

The figure~\ref{fig:nodeGraph} illustrates the use of the chain rule to compute the derivative or to "back-propagate the gradient" as commonly said. Input data consists of variables $a$ and $b$, closer to the leaves (\ie the illumination parameters ${\bf L}$) and output data is $c$, closer to the root (\ie the rendered image ${\bf I}$). The node achieves the function $f$ so that $c = f(a,b)$. Assuming that we know the derivative $\partial {\bf I}/\partial c$ of the image w.r.t. the lower part of the node, we can compute the derivatives $\partial {\bf I}/\partial a$ and $\partial {\bf I}/\partial b$ w.r.t. the upper part of the node if and only if $f$ is differentiable and the partial derivatives $\partial f/\partial a$ and $\partial f/\partial b$ are known. Recursively, ${\bf I}$ is differentiable w.r.t. ${\bf L}$ if and only if each node is differentiable and we can compute the derivatives of their outputs w.r.t. their inputs. In practice, automatic differentiation is used to compute the derivatives of each node.

\subsection{Differentiable Cast Shadows}
\label{sec:differentiable}

Differentiating cast shadows ${\bf S}$ is far from trivial. The most common method for shadow synthesis from a point light is shadow mapping. To test if a point is shadowed, a Z-buffer (the shadow map) is rendered from the light source point of view; then the stored depth $z$ is compared to the distance $d$ from the point to the light source, to evaluate the presence of an occluder. Due to the binary nature of the occlusion test, shadow mapping is not differentiable. To make if differentiable we would have to replace the test by an activation such as $(d, z)\rightarrow\arctan(a.|z - d + b|)$, $b$ being the bias to remove shadow acne and $a$ controlling the slope of the activation function. The steeper the slope, the harder the shadow. The trade-off is the following: slope must be steep enough for the shadow not to be too smooth, but hard shadows will lead to degenerated gradients. Unfortunately we were unable to reach such threshold. 

As a solution to make shadows differentiable, we explored different illumination models. Gruber \etal \cite{gruber_real-time_2012, gruber_efficient_2014, gruber_image-space_2015} propose to encode illumination and a radiance transfer (RT) function (containing visibility) in a basis spherical harmonic (SH). Pixel brightness is then a scalar product between illumination and this RT function so shadows produced by SH are differentiable w.r.t. the light source parameters (9 parameters for 3-band SH). However as highlighted by Okabe \etal \cite{okabe_spherical_2004}, SH tend to recover only low-frequency components, thus is not suited to model point light sources and hard shadows due to their localized nature in the angular domain. Instead they propose a basis of Haar wavelets to model all-frequency illumination. Nevertheless, like SH, Haar wavelets only model distant lighting (illumination is reduced to 2D directions). To cope with non-distant light sources and induced lighting parallax (the shadow cast by an object depends on its position), we keep modeling illumination with 3D point light sources.

To solve the problem of shadow differentiability for non-distant light sources, we perform shadow mapping and we approximate the Jacobian matrix $\partial {\bf S}/\partial {\bf L}$ with finite differences. For instance, six shadow maps are rendered around the current light source position estimation to numerically approximate the Jacobian matrix $\partial {\bf S}/\partial {\bf L}$, that is integrated to the chained derivation as:
\begin{equation}
    \frac{\partial {\bf I}}{\partial {\bf L}} = \frac{\partial {\bf I}}{\partial {\bf I}_d}\frac{\partial {\bf I}_d}{\partial {\bf L}} + \frac{\partial {\bf I}}{\partial {\bf I}_s}\frac{\partial {\bf I}_s}{\partial {\bf L}} + ({\bf I}_d + {\bf I}_s).\frac{\partial {\bf S}}{\partial {\bf L}}.
\end{equation}
Other derivatives are fast-forward: according to (\ref{eq:phong}) they are found by deriving simple differentiable operations like the dot product or the element-wise multiplication. The whole graph is implemented with the automatic differentiation library Pytorch \cite{pytorch}. The energy (\ref{eq:energy1}) is minimized via a gradient descent algorithm.

\section{Experiments and Results}
\label{sec:experiments}

\begin{figure}[t!]
\begin{center}
\begin{tabular}{cc}
\includegraphics[width=0.45\columnwidth]{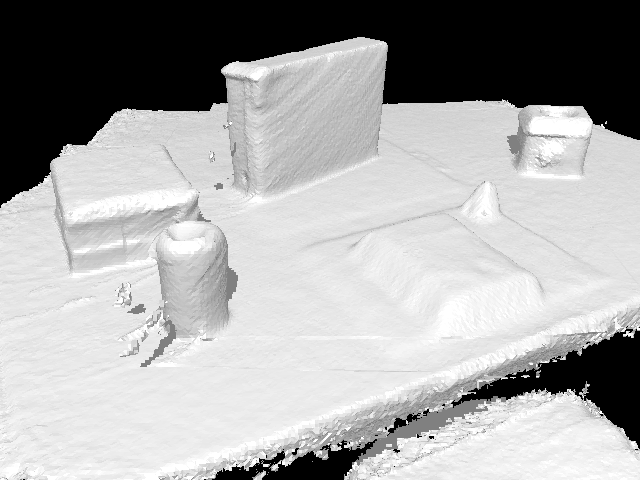} &
\includegraphics[width=0.45\columnwidth]{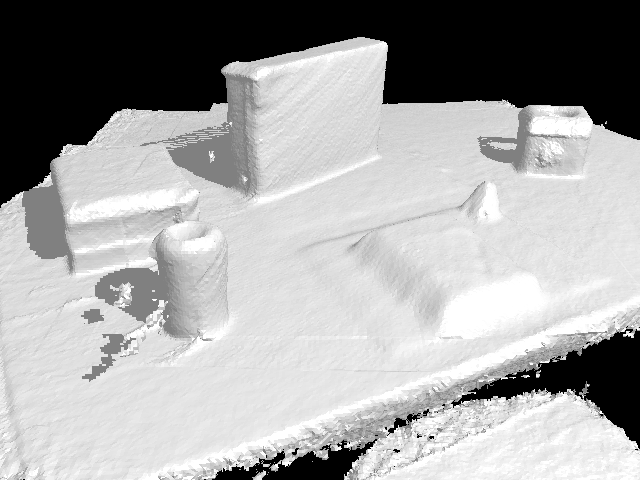} \\
(a) & (b) \\
\includegraphics[width=0.45\columnwidth]{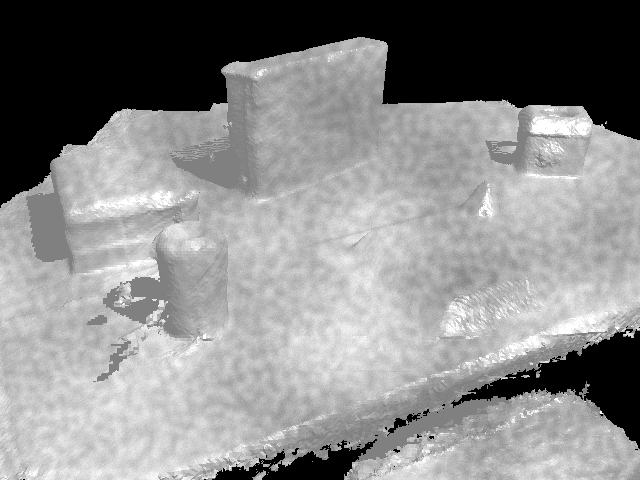} &
\includegraphics[width=0.45\columnwidth]{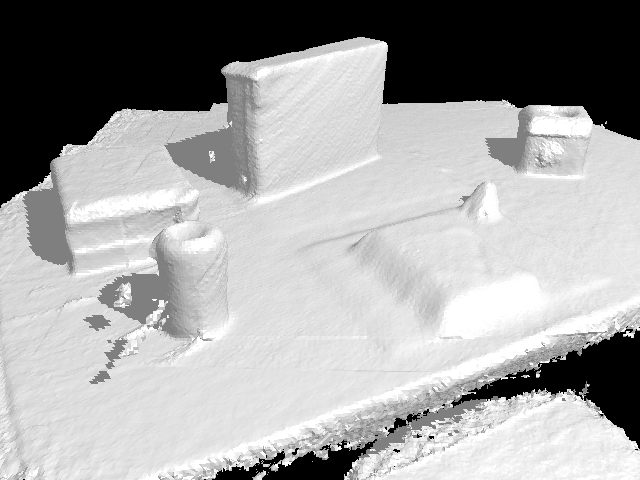} \\
(c) & (d) \\
\end{tabular}
\end{center}
\caption{Example of a robust estimation. Our estimation for an ideal reflectance is outperformed by \cite{neverova_lighting_2012, boom2013point}, but outperforms them in the case of a noisy reflectance (a) Initialization. (b) Final estimation. (c) Final estimation with noisy reflectance. (d) Observed (target) image.}
\label{fig:synth}
\end{figure}

To demonstrate the robustness of the method to noisy reflectance estimation, we test it on a synthetic dataset. The dataset is composed of input images that have been rendered with Unity \cite{unity}, from 3D models of real indoor scenes that where acquired with an RGBD camera. Images are rendered with ideal material parameters: ${\bf k}_d = {\bf k}_s = 1.0$, $\alpha = 10$ and ${\bf I}_a = 0.5$. In total, 9 scenes have been synthesized, with 6 different light source positions each, distributed around the scene to cover very different lighting conditions. For each scene, we aim to estimate the position of a single point light source. Note that we could estimate the color of the light likewise. For the sake of simplicity all images are grayscale and the intensity of the light source in our renderer is set to its true value ${\bf I}_{L} = 0.5$.

\begin{figure*}
\begin{center}
\includegraphics[width=0.90\textwidth]{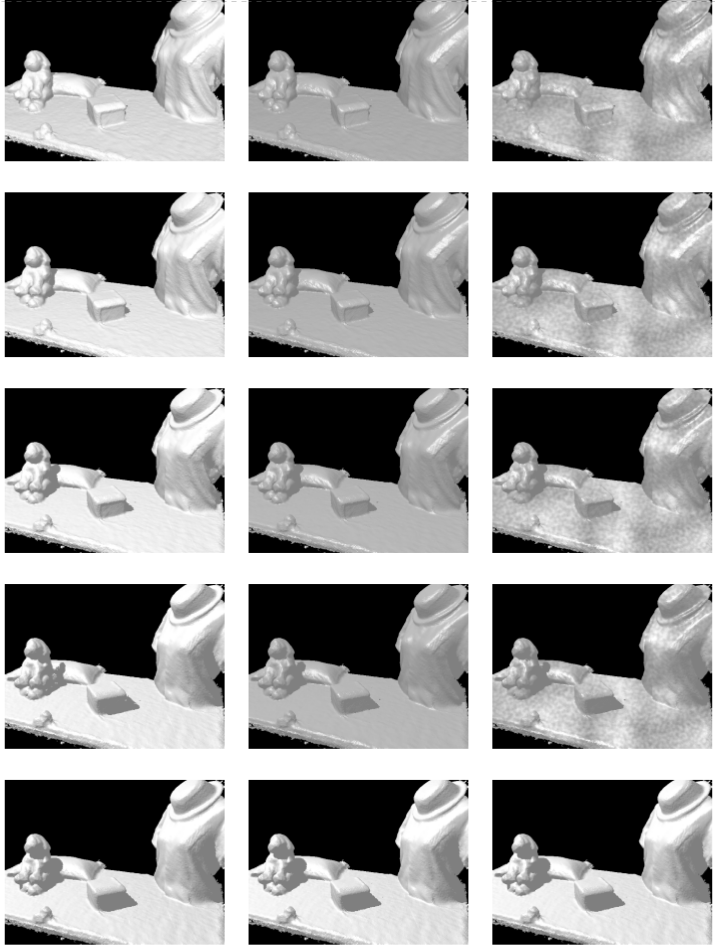}
\end{center}
\caption{An example of estimation of light position 3 in our $5^{th}$ synthetic scene, for different "estimated" reflectances. From left to right: ideal reflectance, no noise ($m = 0.0$) but wrong reflectance value, noisy reflectance ($m = 3.0$). From top to bottom: iteration 0, iteration 20, iteration 40, final iteration (142 for "ideal reflectance", 72 for "$m = 0.0$" and 80 for "$m = 3.0$"). The last row depicts the target image.}
\label{fig:synth_res}
\end{figure*}

For each scene and light source position we compare our approach to state-of-the-art methods \cite{boom2013point, neverova_lighting_2012}. We only focus on their light source estimation, and do not use their intrinsic image decomposition algorithm: we give their models the same material values ${\bf k}_d$, ${\bf k}_s$ and $\alpha$ as ours. The same optimization scheme is used for all methods, for the comparison to be fair: gradient descent for all, with a rate of 0.02 and a stopping condition when the relative energy change falls below $10^{-4}$. Boom \etal's energy \cite{boom2013point} only contains a diffuse term and an ambient term, Neverova \etal \cite{neverova_lighting_2012} append a specular term; we take exactly the same weights as in their paper. To be fair, the same geometry (a depth map and a normal map) is taken for all methods. In addition, we test all three models with different material parameters, to demonstrate their robustness to imperfect reflectance estimations ${\bf k}_d$ and ${\bf k}_s$. At first we try ideal reflectances (figure \ref{fig:synth} (b)) --- same values that were used to generate the scenes --- then we try adding some fractal noise (FBM) to the reflectances in the rendering models (figure \ref{fig:synth} (c)). $m$ denotes the magnitude of the added noise. In the case $m = 0$, no noise is added but the reflectance values are set to 0.5 instead of their true value 1.0. The shininess $\alpha$ is never altered. Figure \ref{fig:synth_res} shows some convergence results for various types of "estimated" reflectances.

\begin{table}
\begin{center}
\begin{tabular}{|l|l|c|c|c|}
\cline{3-5}
\multicolumn{2}{c|}{} & \cite{boom2013point} & \cite{neverova_lighting_2012} & Ours \\
\cline{3-5}
\cline{1-5}
& Average error & 0.0944 & {\bf 0.0720} & 0.0928 \\
no noise & Median error & 0.0792 & {\bf 0.0552} & 0.1136 \\
& Success rate (\%) & 20.37 & {\bf 57.41} & 22.22 \\
\cline{1-5}
& Average error & 0.2600 & 0.2379 & {\bf 0.2100} \\
$m = 0.0$ & Median error & 0.2468 & 0.2266 & {\bf 0.2111} \\
& Success rate (\%) & 5.56 & 14.81 & {\bf 79.63} \\
\cline{1-5}
& Average error & 0.2433 & 0.2281 & {\bf 0.1940} \\
$m = 0.1$ & Median error & 0.2429 & 0.2275 & {\bf 0.2040} \\
& Success rate (\%) & 3.70 & 11.11 & {\bf 85.19} \\
\cline{1-5}
& Average error & 0.2245 & 0.2164 & {\bf 0.1782} \\
$m = 0.2$ & Median error & 0.2362 & 0.2284 & {\bf 0.1911} \\
& Success rate (\%) & 5.56 & 7.41 & {\bf 87.04} \\
\cline{1-5}
& Average error & 0.2015 & 0.1964 & {\bf 0.1678} \\
$m = 0.3$ & Median error & 0.2121 & 0.2089 & {\bf 0.1863} \\
& Success rate (\%) & 11.11 & 12.96 & {\bf 75.93} \\
\cline{1-5}        
\end{tabular}
\end{center}
\caption{Comparison with Boom \etal \cite{boom2013point} and Neverova \etal \cite{neverova_lighting_2012}. Best results (lower light source position error and higher success rate) are in bold.}
\label{tab:results}
\end{table}

\begin{figure}[t!]
\begin{center}
\begin{tabular}{cc}
\includegraphics[width=0.45\columnwidth]{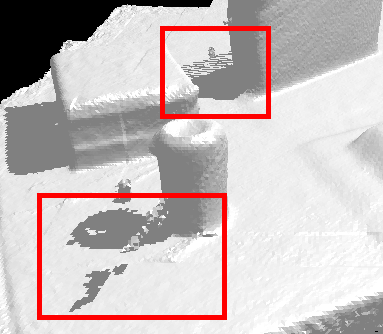} &
\includegraphics[width=0.45\columnwidth]{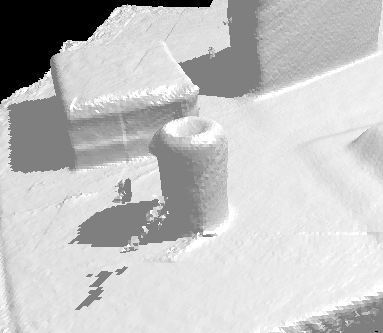} \\
(a) & (b) \\
\end{tabular}
\end{center}
\caption{(a) Final estimation. (b) Target image. Given the limited and noisy depth information, our method cannot produce perfect shadows (red rectangles), and yet accurately estimates illumination.}
\label{fig:fail}
\end{figure}

The table \ref{tab:results} shows the numerical results: the average and median error on the estimation of the light source position and the success rate over the 54 experiments. Convergence is achieved in a minute in average. Our method does not distinguish in the scenario of an ideal material estimation, but systematically achieves better light source estimation in the cases of incorrect reflectance estimation. The figure \ref{fig:synth} shows an example of a case where our method outperforms \cite{neverova_lighting_2012, boom2013point} in the case of a noisy reflectance. We think incorrect shadows are disadvantageous in the case where an image rendered with a simple Phong model already perfectly fits the observed data. Since to be fair the same geometry information is used for all methods, the same depth map (converted to point cloud) is also used to compute the shadow maps. The resulting shadows (red rectangles in figure \ref{fig:fail}) are either incomplete (only the visible depth information is used) or aliased (coarse point-based rendering is performed), which does not serve our method. However we strongly believe that the use of a full mesh would solve the problem and improve our results in the case of an ideal reflectance estimation. In other cases, the use of even imperfect shadows has been proved to be very effective.  

\begin{figure*}
\begin{center}
\begin{tabular}{ccc}
\includegraphics[width=0.58\columnwidth]{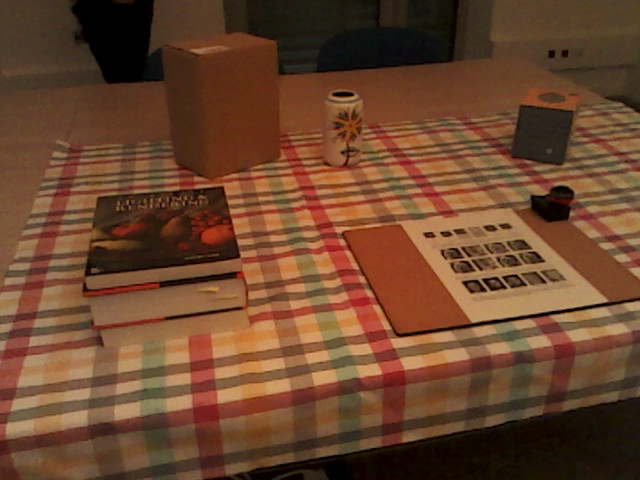} &
\includegraphics[width=0.58\columnwidth]{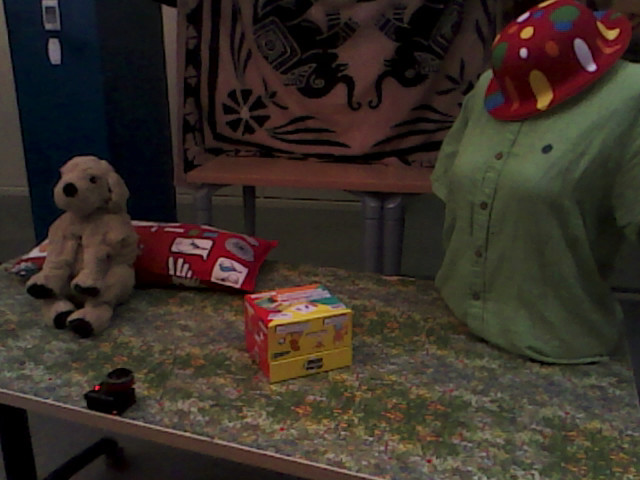} &
\includegraphics[width=0.58\columnwidth]{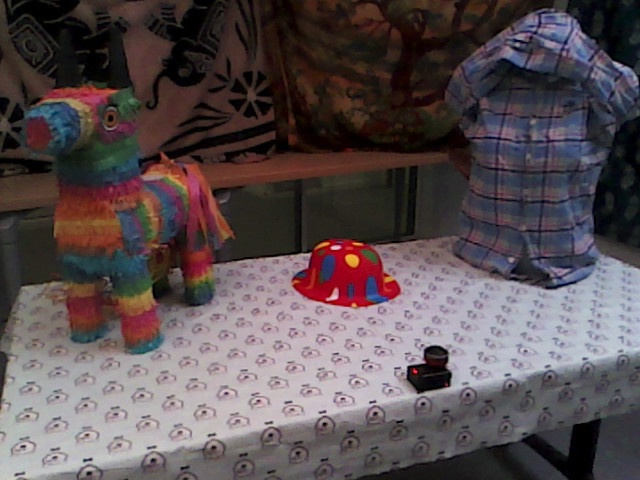} \\
&&\\
\includegraphics[width=0.58\columnwidth]{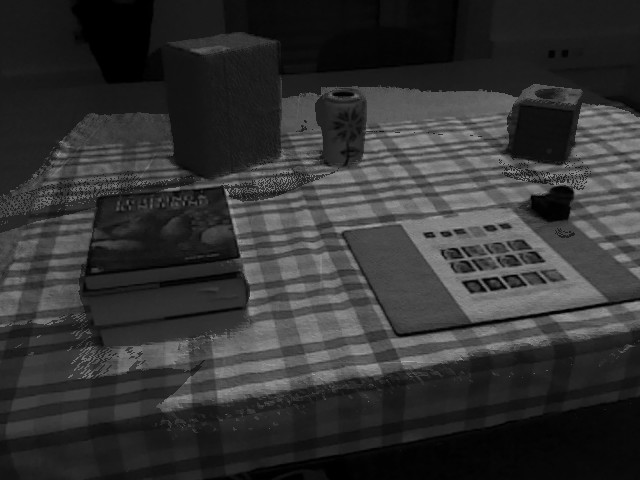} &
\includegraphics[width=0.58\columnwidth]{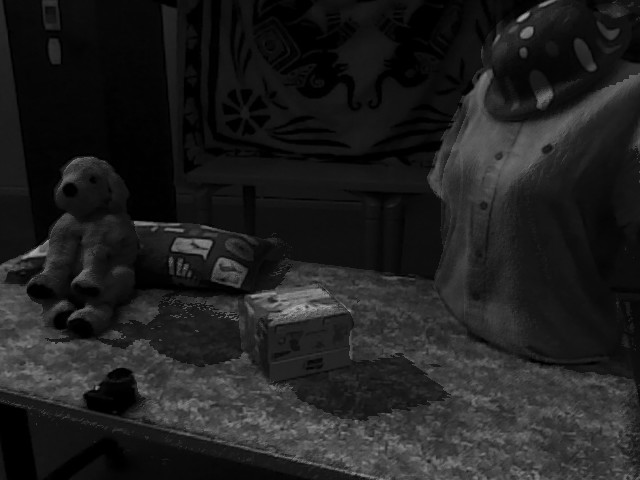} &
\includegraphics[width=0.58\columnwidth]{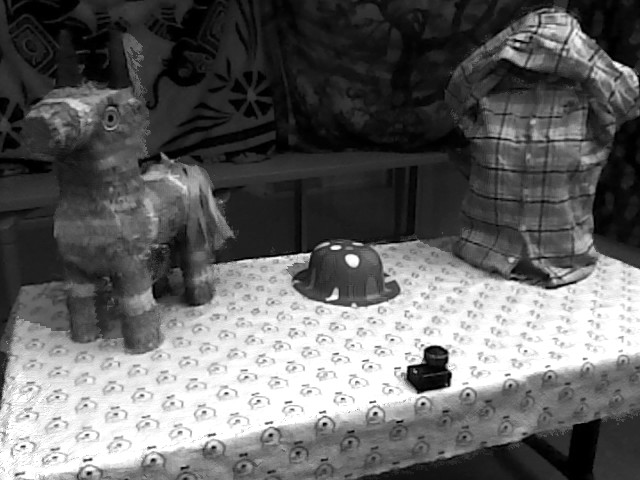} \\
&&\\
\includegraphics[width=0.58\columnwidth]{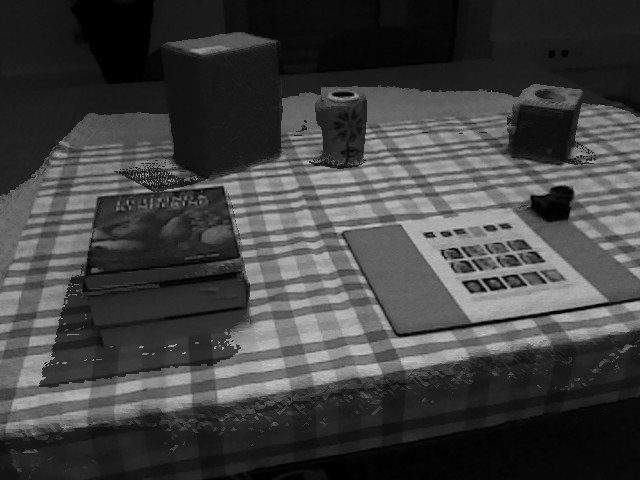} &
\includegraphics[width=0.58\columnwidth]{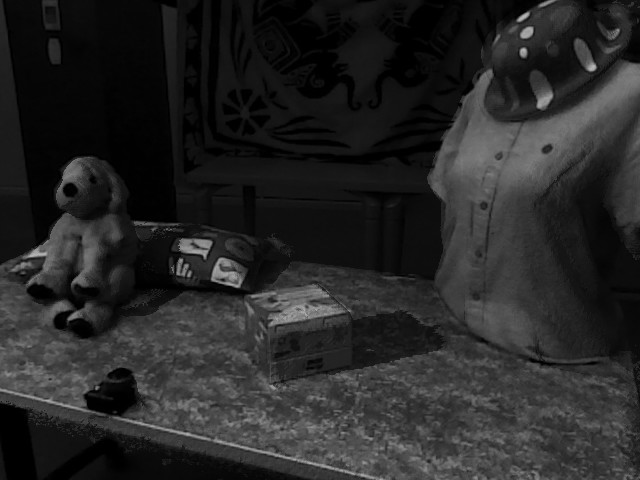} &
\includegraphics[width=0.58\columnwidth]{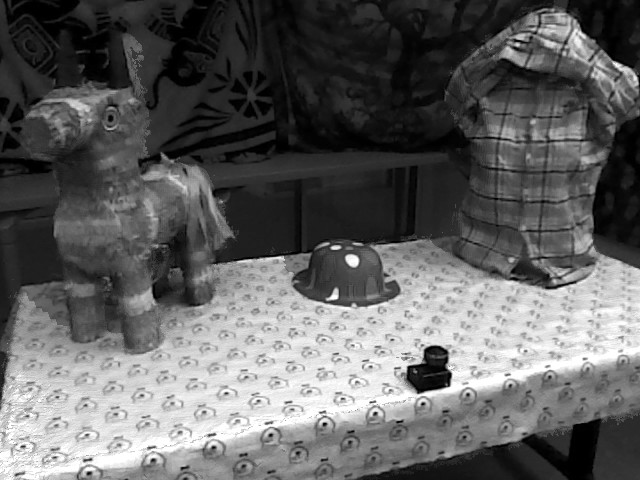} \\
&&\\
\includegraphics[width=0.58\columnwidth]{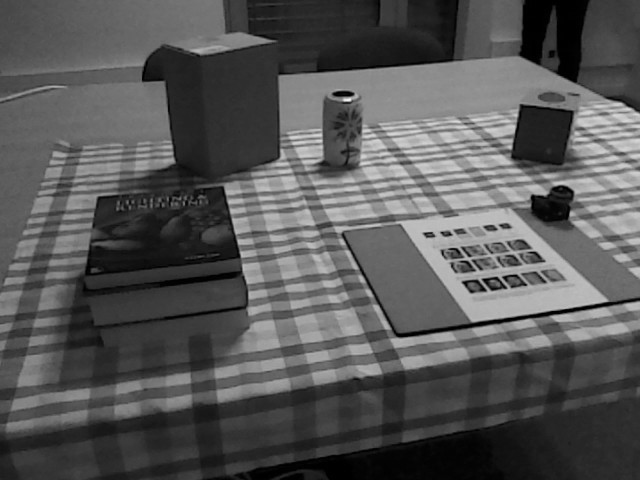} &
\includegraphics[width=0.58\columnwidth]{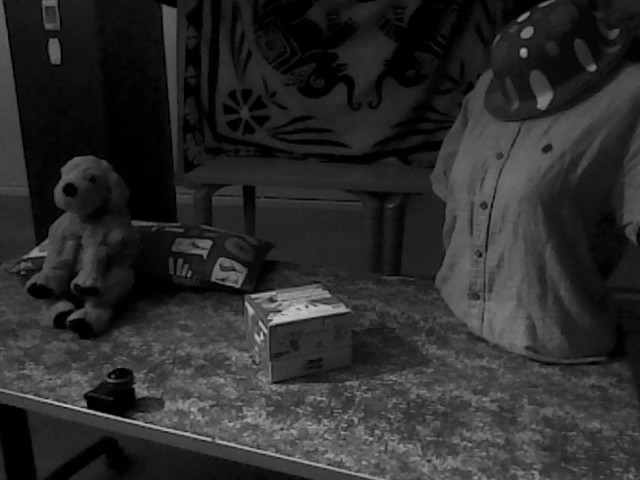} &
\includegraphics[width=0.58\columnwidth]{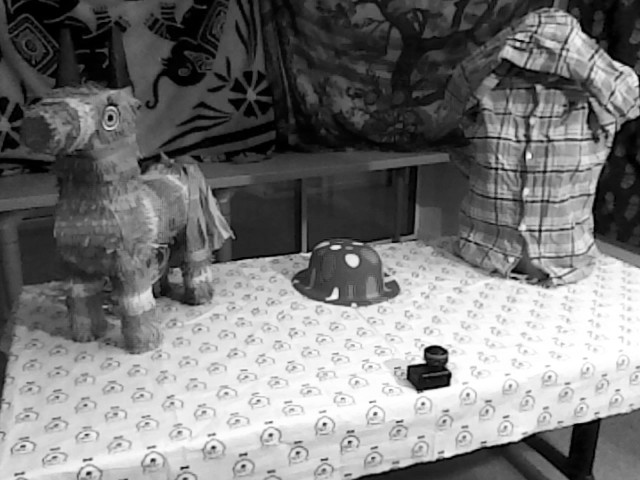} \\
&&\\
\includegraphics[width=0.58\columnwidth]{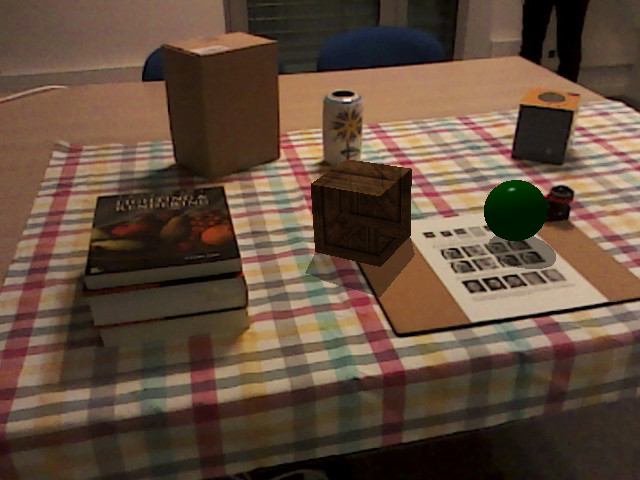} &
\includegraphics[width=0.58\columnwidth]{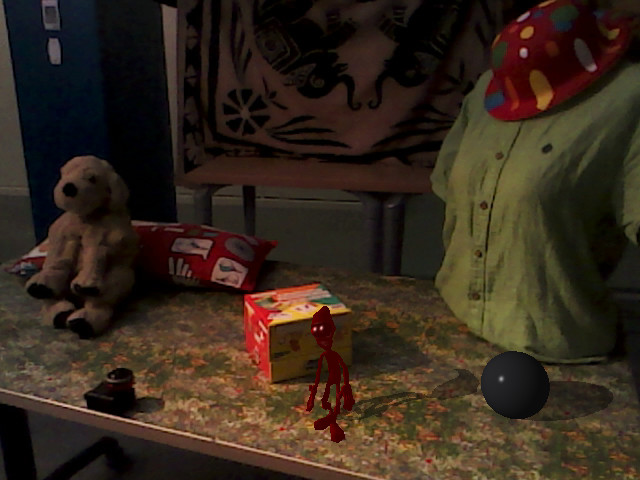} &
\includegraphics[width=0.58\columnwidth]{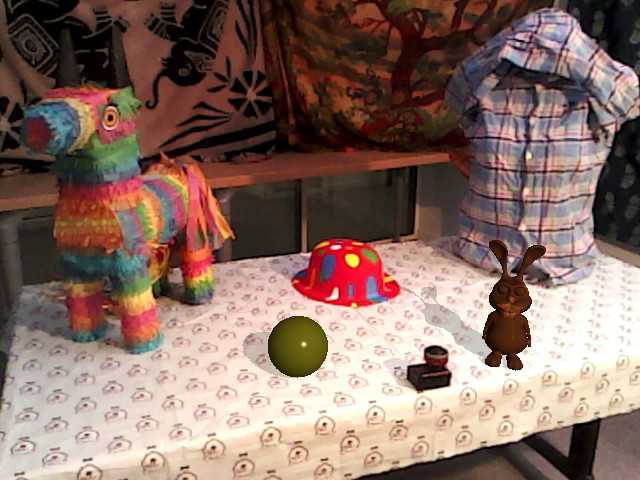} \\
scene 4 & scene 5 & scene 6\\ 
\end{tabular}
\end{center}
\caption{Results on three real scenes. First row: image captured under pseudo-ambient light. Second row: image rendered at initialization. Third row: image rendered at last iteration, for the final estimated light source position. Fourth row: observed (target) image. Last row: insertion of virtual objects with coherent illumination and cast shadow.}
\label{fig:real}
\end{figure*}

A last set of experiments on real data was conducted to demonstrate the capability of our algorithm to find real light source positions. We used the same scenes as before but with real images as input. The figure \ref{fig:real} shows the results on three of these scenes. To estimate the ambient lighting ${\bf I}_a$ and the diffuse reflectance ${\bf k}_d$, we captured each scene under what we call a pseudo-ambient illumination, that we believe minimizes the effect of the illumination and highlights the texture only. The specular reflectance ${\bf k}_s$ is set to 0. The insertion of virtual objects correctly lit by the estimated light source provides a visual clue of the correctness of the estimation. 

\section{Conclusion}
\label{sec:conclusion}

We presented a method to solve the problem of illumination retrieval as a continuous optimization. We perform this optimization via a completely differentiable renderer based on the Blinn-Phong model with cast shadows. We compared our differentiable renderer to state-of-the-art methods and showed it clearly outperforms them in the case of non-ideal reflectance, which is the common practical scenario. We proved that adding a differentiable shadow caster increases the robustness of the estimation. Applied on real scenes, this illumination estimation method provides a plausible light source estimate for the insertion of virtual objects to be consistent with the rest of the scene. In particular, the rendered shadows are coherent with the shadows cast by real objects.

As a future work we intend to include our differentiable renderer in an unsupervised deep learning architecture. We believe it can highly improve the training of a CNN that estimates the illumination of a scene.

{\small
\bibliographystyle{ieee}
\bibliography{biblio}
}

\end{document}